\documentclass[11pt]{article}
\usepackage{acl07}
\usepackage{times}
\usepackage{latexsym}
\usepackage{amssymb,amsmath}
\usepackage{lscape,graphicx}
\usepackage{algorithmic}
\usepackage{algorithm}
\usepackage{rotating}
\usepackage{subfigure}

\newtheorem{definition}{Definition}

\title{Determining the Unithood of Word Sequences using a Probabilistic Approach}

\author{Wilson Wong, Wei Liu and Mohammed Bennamoun\\School of Computer Science and Software Engineering\\University of Western Australia \\Crawley WA 6009 \\  {\{wilson,wei,bennamou\}@csse.uwa.edu.au}}

\begin{document}
\maketitle
\begin{abstract}
Most research related to unithood were conducted as part of a larger effort for the determination of termhood. Consequently, novelties are rare in this small sub-field of term extraction. In addition, existing work were mostly empirically motivated and derived. We propose a new probabilistically-derived measure, independent of any influences of termhood, that provides dedicated measures to gather linguistic evidence from parsed text and statistical evidence from Google search engine for the measurement of unithood. Our comparative study using $1,825$ test cases against an existing empirically-derived function revealed an improvement in terms of precision, recall and accuracy.
\end{abstract}

\section{Introduction}
\emph{Automatic term recognition}, also referred to as \emph{term extraction} or \emph{terminology mining}, is the process of extracting lexical units from text and filtering them for the purpose of identifying terms which characterise certain domains of interest. This process involves the determination of two factors: \emph{unithood} and \emph{termhood}. Unithood concerns with whether or not a sequence of words should be combined to form a more stable lexical unit. On the other hand, termhood measures the degree to which these stable lexical units are related to domain-specific concepts. Unithood is only relevant to \emph{complex terms} (i.e. multi-word terms) while termhood \cite{wong_et_al_2007e} deals with both \emph{simple terms} (i.e. single-word terms) and complex terms. Recent reviews by \cite{wong_et_al_2007b} show that existing research on unithood are mostly carried out as a prerequisite to the determination of termhood. As a result, there is only a small number of existing measures dedicated to determining unithood. Besides the lack of dedicated attention in this sub-field of term extraction, the existing measures are usually derived from term or document frequency, and are modified as per need. As such, the significance of the different weights that compose the measures usually assume an empirical viewpoint. Obviously, such methods are at most inspired by, but not derived from formal models \cite{kageura_umino_1996}.

The three objectives of this paper are (1) to separate the measurement of unithood from the determination of termhood, (2) to devise a probabilistically-derived measure which requires only one threshold for determining the unithood of word sequences using non-static textual resources, and (3) to demonstrate the superior performance of the new probabilistically-derived measure against existing empirical measures. In regards to the first objective, we will derive our probabilistic measure free from any influence of termhood determination. Following this, our unithood measure will be an independent tool that is applicable not only to term extraction, but many other tasks in information extraction and text mining. Concerning the second objective, we will devise our new measure, known as the \emph{Odds of Unithood $(OU)$}, which are derived using Bayes Theorem and founded on a few elementary probabilities. The probabilities are estimated using Google page counts in an attempt to eliminate problems related to the use of static corpora. Moreover, only one threshold, namely, $OU_T$ is required to control the functioning of $OU$. Regarding the third objective, we will compare our new $OU$ against an existing empirically-derived measure called \emph{Unithood $(UH)$} \cite{wong_et_al_2007b} in terms of their precision, recall and accuracy.

In Section \ref{related_work}, we provide a brief review on some of existing techniques for measuring unithood. In Section \ref{our_approach}, we present our new probabilistic approach, the measures involved, and the theoretical and intuitive justification behind every aspect of our measures. In Section \ref{evaluation}, we summarize some findings from our evaluations. Finally, we conclude this paper with an outlook to future work in Section \ref{conclusion}.

\section{Related Works}\label{related_work}

Some of the most common measures of unithood include pointwise \emph{mutual information (MI)} \cite{church_hanks_1990} and \emph{log-likelihood ratio} \cite{dunning_1994}. In mutual information, the co-occurrence frequencies of the constituents of complex terms are utilised to measure their dependency. The mutual information for two words $a$ and $b$ is defined as:
\begin{eqnarray}
\label{mi}
MI(a,b) = \log_2{\frac{p(a,b)}{p(a)p(b)}}
\end{eqnarray}
where $p(a)$ and $p(b)$ are the probabilities of occurrence of $a$ and $b$. Many measures that apply statistical techniques assuming strict normal distribution, and independence between the word occurrences \cite{franz_1997} do not fare well. For handling extremely uncommon words or small sized corpus, \emph{log-likelihood ratio} delivers the best precision \cite{kurz_xu_2002c}. Log-likelihood ratio attempts to quantify how much more likely one pair of words is to occur compared to the others. Despite its potential, \emph{``How to apply this statistic measure to quantify structural dependency of a word sequence remains an interesting issue to explore.''} \cite{kit_2002}. \cite{seretan_et_al_2004} tested mutual information, log-likelihood ratio and t-tests to examine the use of results from web search engines for determining the collocational strength of word pairs. However, no performance results were presented. 

\cite{wong_et_al_2007b} presented a hybrid approach inspired by mutual information in Equation \ref{mi}, and \emph{C-value} in Equation \ref{cvalue}. The authors employ Google page counts for the computation of statistical evidences to replace the use of frequencies obtained from static corpora. Using the page counts, the authors proposed a function known as \emph{Unithood (UH)} for determining the mergeability of two lexical units $a_x$ and $a_y$ to produce a stable sequence of words $s$. The word sequences are organised as a set $W=\{s,a_x,a_y\}$ where $s=a_{x}b a_{y}$ is a term candidate, $b$ can be any preposition, the coordinating conjunction \emph{``and"} or an empty string, and $a_x$ and $a_y$ can either be noun phrases in the form $Adj^*N+$ or another $s$ (i.e. defining a new $s$ in terms of other $s$). The authors define $UH$ as:
\begin{equation} 
\label{to_merge_or_not}
UH(a_x,a_y) = 
\begin{cases}
1 & \text{if $(MI(a_x,a_y) > MI^+ )$ $\lor$} \\
	&\quad \text{$(MI^+ \geq MI(a_x,a_y)$}\\
	&\quad \text{$\geq MI^-\land$}\\
	&\quad \text{$ID(a_x,s) \geq ID_T$  $\land$}\\
	&\quad \text{$ID(a_y,s) \geq ID_T$  $\land$}\\
	&\quad \text{$IDR^+ \geq IDR(a_x,a_y)$}\\
	&\quad \text{$\geq IDR^-)$}\\
0 &\text{otherwise}
\end{cases}
\end{equation}
where $MI^+$, $MI^-$, $ID_T$, $IDR^+$ and $IDR^-$ are thresholds for determining mergeability decisions, and $MI(a_x,a_y)$ is the mutual information between $a_x$ and $a_y$, while $ID(a_x,s)$, $ID(a_y,s)$ and $IDR(a_x,a_y)$ are measures of lexical independence of $a_x$ and $a_y$ from $s$. For brevity, let $z$ be either $a_x$ or $a_y$, and the independence measure $ID(z,s)$ is then defined as:
\begin{equation} 
ID(z,s) = 
\begin{cases}
\log_{10}(n_{z} - n_s)&\text{if($n_{z} > n_s$)}\nonumber\\
0											&\text{otherwise}\nonumber
\end{cases}
\end{equation} 
where $n_{z}$ and $n_s$ is the Google page count for $z$ and $s$ respectively. On the other hand, $IDR(a_x,a_y)=\frac{ID(a_x,s)}{ID(a_y,s)}$. Intuitively, $UH(a_x,a_y)$ states that the two lexical units $a_x$ and $a_y$ can only be merged in two cases, namely, 1) if $a_x$ and $a_y$ has extremely high mutual information (i.e. higher than a certain threshold $MI^+$), or 2) if $a_x$ and $a_y$ achieve average mutual information (i.e. within the acceptable range of two thresholds $MI^+$ and $MI^-$) due to both of their extremely high independence (i.e. higher than the threshold $ID_T$) from $s$.

\cite{frantzi_1997} proposed a measure known as \emph{Cvalue} for extracting complex terms. The measure is based upon the claim that a substring of a term candidate is a candidate itself given that it demonstrates adequate independence from the longer version it appears in. For example, \emph{``E. coli food poisoning''}, \emph{``E. coli''} and \emph{``food poisoning''} are acceptable as valid complex term candidates. However, \emph{``E. coli food''} is not. Therefore, some measures are required to gauge the strength of word combinations to decide whether two word sequences should be merged or not. Given a word sequence $a$ to be examined for unithood, the \emph{Cvalue} is defined as:
\begin{equation} 
\label{cvalue}
Cvalue(a) = 
\begin{cases}
\log_2|a|f_a 																			& \text{if $|a| = g$}\\
\log_2|a|(f_a-\frac{\sum_{l\in L_a}f_l}{|L_a|})& \text{otherwise}
\end{cases}
\end{equation} 
where $|a|$ is the number of words in $a$, $L_a$ is the set of longer term candidates that contain $a$, $g$ is the longest n-gram considered, $f_a$ is the frequency of occurrence of $a$, and $a \notin L_a$. While certain researchers \cite{kit_2002} consider \emph{Cvalue} as a termhood measure, others \cite{nakagawa_mori_2002} accept it as a measure for unithood. One can observe that longer candidates tend to gain higher weights due to the inclusion of $\log_{2}|a|$ in Equation \ref{cvalue}. In addition, the weights computed using Equation \ref{cvalue} are purely dependent on the frequency of $a$.

\section{A Probabilistically-derived Measure for Unithood Determination}\label{our_approach}
We propose a probabilistically-derived measure for determining the unithood of word pairs (i.e. potential term candidates) extracted using the head-driven left-right filter \cite{wong_2005,wong_et_al_2007b} and Stanford Parser \cite{klein_manning_2003}. These word pairs will appear in the form of $(a_x,a_y) \in A$ with $a_x$ and $a_y$ located immediately next to each other (i.e. $x+1=y$), or separated by a preposition or coordinating conjunction \emph{``and''} (i.e. $x+2=y$). Obviously, $a_x$ has to appear before $a_y$ in the sentence or in other words, $x<y$ for all pairs where $x$ and $y$ are the word offsets produced by the Stanford Parser. The pairs in $A$ will remain as potential term candidates until their unithood have been examined. Once the unithood of the pairs in $A$ have been determined, they will be referred to as \emph{term candidates}. Formally, the unithood of any two lexical units $a_x$ and $a_y$ can be defined as
\begin{definition}\label{problem} 
\emph{The unithood of two lexical units is the} ``degree of strength or stability of syntagmatic combinations and collocations" \cite{kageura_umino_1996} \emph{between them.}
\end{definition}
It is obvious that the problem of measuring the unithood of any pair of words is the determination of their \emph{``degree"} of collocational strength as mentioned in Definition \ref{problem}. In practical terms, the \emph{``degree"} mentioned above will provide us with a way to determine if the units $a_x$ and $a_y$ should be combined to form $s$, or left alone as separate units. The collocational strength of $a_x$ and $a_y$ that exceeds a certain threshold will demonstrate to us that $s$ is able to form a stable unit and hence, a better term candidate than $a_x$ and $a_y$ separated. It is worth pointing that the size (i.e. number of words) of $a_x$ and $a_y$ is not limited to $1$. For example, we can have \emph{$a_x$=``National Institute''}, \emph{$b$=``of''} and \emph{$a_y$=``Allergy and Infectious Diseases''}. In addition, the size of $a_x$ and $a_y$ has no effect on the determination of their unithood using our approach.

As we have discussed in Section \ref{related_work}, most of the conventional practices employ frequency of occurrence from local corpora, and some statistical tests or information-theoretic measures to determine the coupling strength between elements in $W = \{s,a_x,a_y\}$. Two of the main problems associated with such approaches are: 
\begin{itemize}
\item Data sparseness is a problem that is well-documented by many researchers \cite{keller_et_al_2002}. It is inherent to the use of local corpora that can lead to poor estimation of parameters or weights; and
\item Assumption of independence and normality of word distribution are two of the many problems in language modelling \cite{franz_1997}. While the independence assumption reduces text to simply a bag of words, the assumption of normal distribution of words will often lead to incorrect conclusions during statistical tests.
\end{itemize}
As a general solution, we innovatively employ results from web search engines for use in a probabilistic framework for measuring unithood. 

As an attempt to address the first problem, we utilise page counts by Google for estimating the probability of occurrences of the lexical units in $W$. We consider the World Wide Web as a large general corpus and the Google search engine as a gateway for accessing the documents in the general corpus. Our choice of using Google to obtain the page count was merely motivated by its extensive coverage. In fact, it is possible to employ any search engines on the World Wide Web for this research. As for the second issue, we attempt to address the problem of determining the degree of collocational strength in terms of probabilities estimated using Google page count. We begin by defining the sample space, $N$ as the set of all documents indexed by Google search engine. We can estimate the index size of Google, $|N|$ using function words as predictors. Function words such as \emph{``a"}, \emph{``is"} and \emph{``with"}, as opposed to content words, appear with frequencies that are relatively stable over many different genres. Next, we perform random draws (i.e. trial) of documents from $N$. For each lexical unit $w \in W$, there will be a corresponding set of outcomes (i.e. events) from the draw. There will be three basic sets which are of interest to us:

\begin{definition}\label{event}
Basic events corresponding to each $w\in W$:
\begin{itemize}
\item $X$ is the event that $a_x$ occurs in the document
\item $Y$ is the event that $a_y$ occurs in the document
\item $S$ is the event that $s$ occurs in the document
\end{itemize}
\end{definition}
It should be obvious to the readers that since the documents in $S$ have to contain all two units $a_x$ and $a_y$, $S$ is a subset of $X \cap Y$ or $S \subseteq X \cap Y$. It is worth noting that even though $S\subseteq X\cap Y$, it is highly unlikely that $S = X \cap Y$ since the two portions $a_x$ and $a_y$ may exist in the same document without being conjoined by $b$. Next, subscribing to the frequency interpretation of probability, we can obtain the probability of the events in Definition \ref{event} in terms of Google page count:
\begin{align}
\label{basic_probability}
P(X) & = \frac{n_x}{|N|}\\\nonumber
P(Y) & = \frac{n_y}{|N|}\\\nonumber
P(S) & = \frac{n_s}{|N|}\nonumber
\end{align}
where $n_x$, $n_y$ and $n_s$ is the page count returned as the result of Google search using the term \emph{[+``$a_x$"]}, \emph{[+``$a_y$"]} and \emph{[+``$s$"]}, respectively. The pair of quotes that encapsulates the search terms is the \emph{phrase} operator, while the character \emph{``+"} is the \emph{required} operator supported by the Google search engine. As discussed earlier, the independence assumption required by certain information-theoretic measures and other Bayesian approaches may not always be valid, especially when we are dealing with linguistics. As such, $P(X \cap Y) \ne P(X)P(Y)$ since the occurrences of $a_x$ and $a_y$ in documents are inevitably governed by some hidden variables and hence, not independent. Following this, we define the probabilities for two new sets which result from applying some set operations on the basic events in Definition \ref{event}:
\begin{align}
\label{derived_probability}
P(X \cap Y) & = \frac{n_{xy}}{|N|}\\\nonumber
P(X \cap Y \setminus S) & = P(X\cap Y)-P(S)\nonumber
\end{align}
where $n_{xy}$ is the page count returned by Google for the search using \emph{[+``$a_x$" +``$a_y$"]}. Defining $P(X\cap Y)$ in terms of observable page counts, rather than a combination of two independent events will allow us to avoid any unnecessary assumption of independence.

Next, referring back to our main problem discussed in Definition \ref{problem}, we are required to estimate the strength of collocation of the two units $a_x$ and $a_y$. Since there is no standard metric for such measurement, we propose to address the problem from a probabilistic perspective. We introduce the probability that $s$ is a stable lexical unit given the evidence $s$ possesses:
\begin{definition}\label{pu}
Probability of unithood:
\begin{align}
P(U|E)=\frac{P(E|U)P(U)}{P(E)}\nonumber
\end{align}
\end{definition}
where $U$ is the event that $s$ is a stable lexical unit and $E$ is the evidences belonging to $s$. $P(U|E)$ is the posterior probability that $s$ is a stable unit given the evidence $E$. $P(U)$ is the prior probability that $s$ is a unit without any evidence, and $P(E)$ is the prior probability of evidences held by $s$. As we shall see later, these two prior probabilities will be immaterial in the final computation of unithood. Since $s$ can either be a stable unit or not, we can state that,
\begin{align}\label{unitandnotunit}
P(\bar{U}|E) = 1 - P(U|E)
\end{align}
where $\bar{U}$ is the event that $s$ is not a stable lexical unit. Since $Odds = P/(1-P)$, we multiply both sides of Definition \ref{pu} by $(1-P(U|E))^{-1}$ to obtain,
\begin{align}\label{odds_1}
\frac{P(U|E)}{1-P(U|E)}=\frac{P(E|U)P(U)}{P(E)(1-P(U|E))}
\end{align}
By substituting Equation \ref{unitandnotunit} in Equation \ref{odds_1} and later, applying the multiplication rule $P(\bar{U}|E)P(E)=P(E|\bar{U})P(\bar{U})$ to it, we will obtain:
\begin{align}\label{odds_2}
\frac{P(U|E)}{P(\bar{U}|E)}=\frac{P(E|U)P(U)}{P(E|\bar{U})P(\bar{U})}
\end{align}
We proceed to take the log of the odds in Equation \ref{odds_2} (i.e. \emph{logit}) to get:
\begin{align}\label{odds_3}
\log{\frac{P(E|U)}{P(E|\bar{U})}} = \log{\frac{P(U|E)}{P(\bar{U}|E)}} - \log{\frac{P(U)}{P(\bar{U})}}
\end{align}
While it is obvious that certain words tend to co-occur more frequently than others (i.e. idioms and collocations), such phenomena are largely arbitrary \cite{smadja_1993}. This makes the task of deciding on what constitutes an acceptable collocation difficult. The only way to objectively identify stable lexical units is through observations in samples of the language (e.g. text corpus) \cite{mckeown_radev_2000}. In other words, assigning the apriori probability of collocational strength without empirical evidence is both subjective and difficult. As such, we are left with the option to assume that the probability of $s$ being a stable unit and not being a stable unit without evidence is the same (i.e. $P(U)=P(\bar{U})=0.5$). As a result, the second term in Equation \ref{odds_3} evaluates to $0$:
\begin{align}\label{odds_4}
\log{\frac{P(U|E)}{P(\bar{U}|E)}} = \log{\frac{P(E|U)}{P(E|\bar{U})}}
\end{align}
We introduce a new measure for determining the odds of $s$ being a stable unit known as \emph{Odds of Unithood (OU)}:
\begin{definition}\label{ou_1}Odds of unithood
\begin{align}\nonumber
OU(s) = \log{\frac{P(E|U)}{P(E|\bar{U})}}
\end{align}
\end{definition}
Assuming that the evidences in $E$ are independent of one another, we can evaluate $OU(s)$ in terms of:
\begin{align}\label{ou_2}
OU(s) & = \log{\frac{\prod_{i}P(e_i|U)}{\prod_{i}P(e_i|\bar{U})}}\\
		& =\sum_{i}\log{\frac{P(e_i|U)}{P(e_i|\bar{U})}}\nonumber
\end{align}
where $e_i$ are individual evidences possessed by $s$.

\begin{figure}[t]
  \centering    
\subfigure[The area with darker shade is the set $X\cap Y\setminus S$. Computing the ratio of $P(S)$ and the probability of this area will give us the first evidence.]{\label{setsA}\includegraphics[width=1.5in]{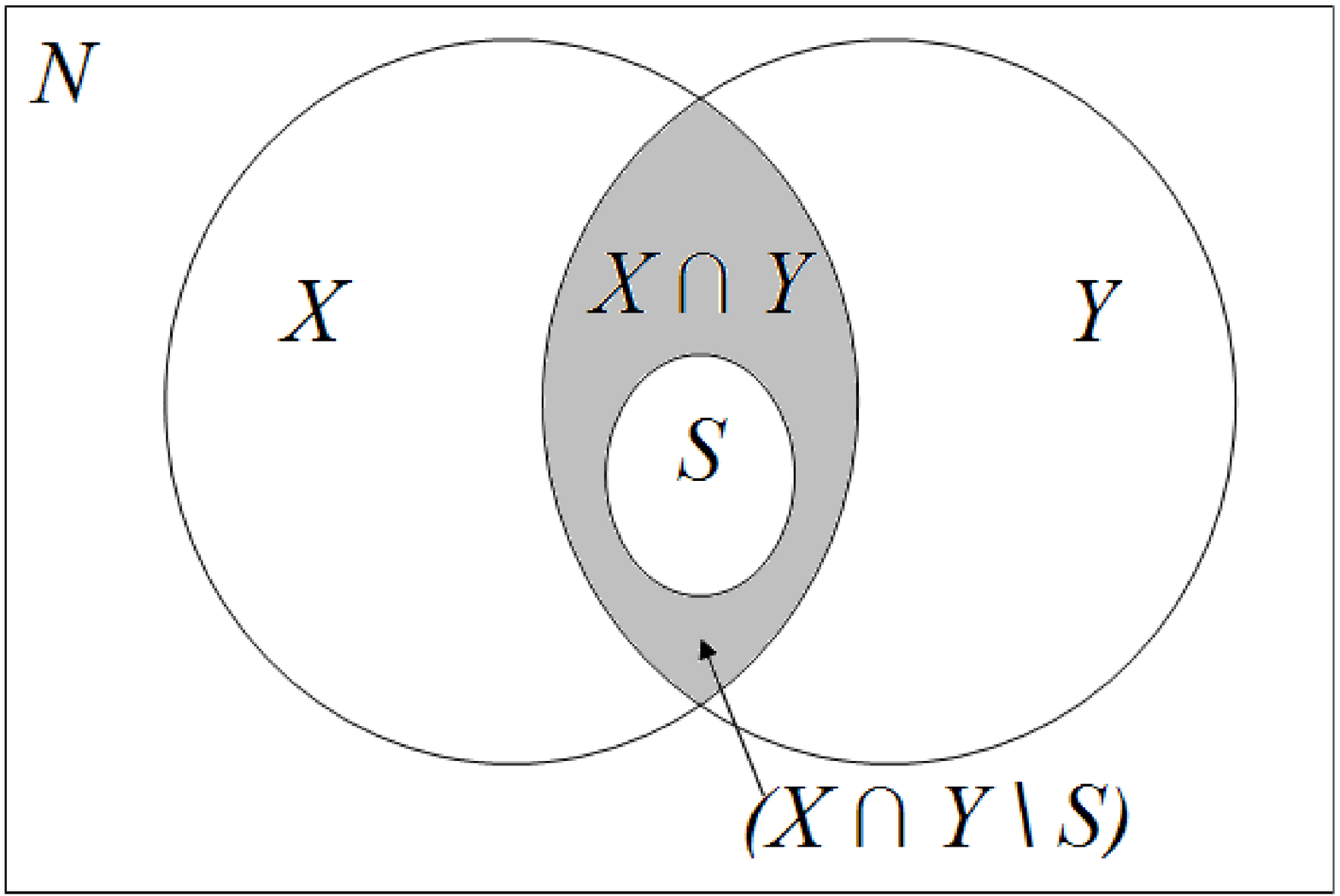}}
\subfigure[The area with darker shade is the set $S^{\prime}$. Computing the ratio of $P(S)$ and the probability of this area (i.e. $P(S^{\prime})=1-P(S)$) will give us the second evidence.]{\label{setsB}\includegraphics[width=1.5in]{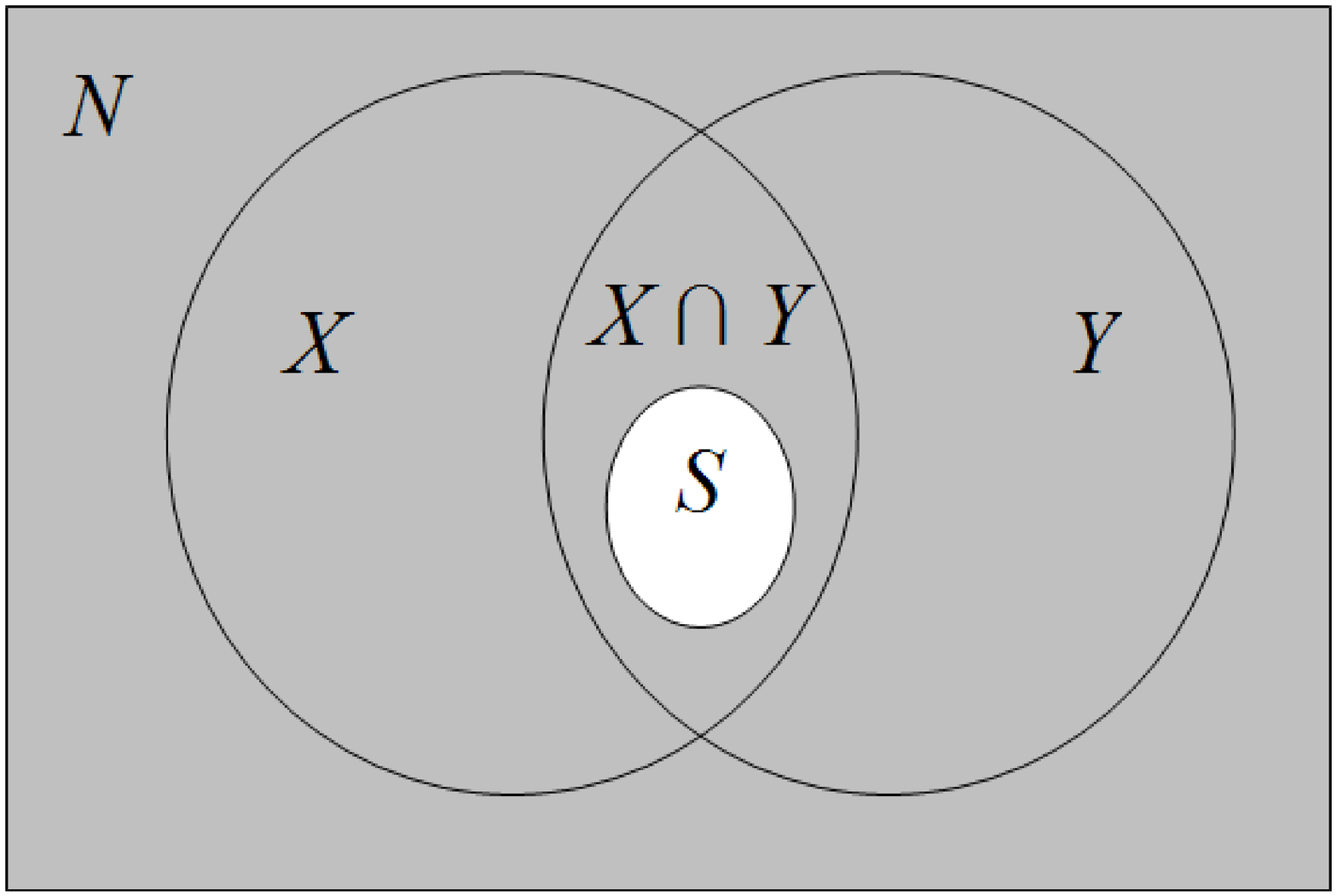}}
  \caption{The probability of the areas with darker shade are the denominators required by the evidences $e_1$ and $e_2$ for the estimation of $OU(s)$.}
\end{figure}

With the introduction of Definition \ref{ou_1}, we can examine the degree of collocational strength of $a_x$ and $a_y$ in forming $s$, mentioned in Definition \ref{problem} in terms of $OU(s)$. With the base of the log in Definition \ref{ou_1} more than 1, the upper and lower bound of $OU(s)$ would be $+\infty$ and $-\infty$, respectively. $OU(s)=+\infty$ and $OU(s)=-\infty$ corresponds to the highest and the lowest degree of stability of the two units $a_x$ and $a_y$ appearing as $s$, respectively. A high\footnote{A subjective issue that may be determined using a threshold} $OU(s)$ would indicate the suitability for the two units $a_x$ and $a_y$ to be merged to form $s$. Ultimately, we have reduced the vague problem of the determination of unithood introduced in Definition \ref{problem} into a practical and computable solution in Definition \ref{ou_1}. The evidences that we propose to employ for determining unithood are based on the occurrences of $s$, or the event $S$ if the readers recall from Definition \ref{event}. We are interested in two types of occurrences of $s$, namely, the occurrence of $s$ given that $a_x$ and $a_y$ have already occurred or $X\cap Y$, and the occurrence of $s$ as it is in our sample space, $N$. We refer to the first evidence $e_1$ as \emph{local occurrence}, while the second one $e_2$ as \emph{global occurrence}. We will discuss the intuitive justification behind each type of occurrences. Each evidence $e_i$ captures the occurrences of $s$ within a different confinement. We will estimate these evidences in terms of the elementary probabilities already defined in Equations \ref{basic_probability} and \ref{derived_probability}. 

The first evidence $e_1$ captures the probability of occurrences of $s$ within the confinement of $a_x$ and $a_y$ or $X\cap Y$. As such, $P(e_1|U)$ can be interpreted as the probability of $s$ occurring within $X\cap Y$ as a stable unit or $P(S|X\cap Y)$. On the other hand, $P(e_1|\bar{U})$ captures the probability of $s$ occurring in $X\cap Y$ not as a unit. In other words, $P(e_1|\bar{U})$ is the probability of $s$ not occurring in $X\cap Y$, or equivalently, equal to $P((X\cap Y\setminus S)|(X\cap Y))$. The set $X\cap Y\setminus S$ is shown as the area with darker shade in Figure \ref{setsA}. Let us define the odds based on the first evidence as:
\begin{align}\label{ol}
O_L = \frac{P(e_1|U)}{P(e_1|\bar{U})}
\end{align}
Substituting $P(e_1|U)=P(S|X\cap Y)$ and $P(e_1|\bar{U})=P((X\cap Y\setminus S)|(X\cap Y))$ into Equation \ref{ol} will give us:
\begin{align}\nonumber
O_L & = \frac{P(S|X\cap Y)}{P((X\cap Y\setminus S)|(X\cap Y))}\\\nonumber
& = \frac{P(S\cap (X\cap Y))}{P(X\cap Y)}\frac{P(X\cap Y)}{P((X\cap Y\setminus S) \cap (X\cap Y))}\\\nonumber
& = \frac{P(S\cap (X\cap Y))}{P((X\cap Y\setminus S) \cap (X\cap Y))}\nonumber
\end{align}
and since $S\subseteq (X\cap Y)$ and $(X\cap Y \setminus S)\subseteq (X\cap Y)$,
\begin{align}\nonumber
O_L & = \frac{P(S)}{P(X\cap Y\setminus S)} & if(P(X\cap Y\setminus S) \ne 0)
\end{align}
and $O_L=1$ if $P(X\cap Y\setminus S)=0$.

The second evidence $e_2$ captures the probability of occurrences of $s$ without confinement. If $s$ is a stable unit, then its probability of occurrence in the sample space would simply be $P(S)$. On the other hand, if $s$ occurs not as a unit, then its probability of non-occurrence is $1-P(S)$. The complement of $S$, which is the set $S^{\prime}$ is shown as the area with darker shade in Figure \ref{setsB}. Let us define the odds based on the second evidence as:
\begin{align}\label{og}
O_G = \frac{P(e_2|U)}{P(e_2|\bar{U})}
\end{align}
Substituting $P(e_2|U)=P(S)$ and $P(e_2|\bar{U})=1-P(S)$ into Equation \ref{og} will give us:
\begin{align}\nonumber
O_G & = \frac{P(S)}{1-P(S)}
\end{align}

Intuitively, the first evidence attempts to capture the extent to which the existence of the two lexical units $a_x$ and $a_y$ is attributable to $s$. Referring back to $O_L$, whenever the denominator $P(X\cap Y \setminus S)$ becomes less than $P(S)$, we can deduce that $a_x$ and $a_y$ actually exist together as $s$ more than in other forms. At one extreme when $P(X\cap Y \setminus S)=0$, we can conclude that the co-occurrence of $a_x$ and $a_y$ is exclusively for $s$. As such, we can also refer to $O_L$ as a measure of exclusivity for the use of $a_x$ and $a_y$ with respect to $s$. This first evidence is a good indication for the unithood of $s$ since the more the existence of $a_x$ and $a_y$ is attributed to $s$, the stronger the collocational strength of $s$ becomes. Concerning the second evidence, $O_G$ attempts to capture the extent to which $s$ occurs in general usage (i.e. World Wide Web). We can consider $O_G$ as a measure of pervasiveness for the use of $s$. As $s$ becomes more widely used in text, the numerator in $O_G$ will increase. This provides a good indication on the unithood of $s$ since the more $s$ appears in usage, the likelier it becomes that $s$ is a stable unit instead of an occurrence by chance when $a_x$ and $a_y$ are located next to each other. As a result, the derivation of $OU(s)$ using $O_L$ and $O_G$ will ensure a comprehensive way of determining unithood. 

Finally, expanding $OU(s)$ in Equation \ref{ou_2} using Equations \ref{ol} and \ref{og} will give us:
\begin{align}\label{ou_3}
OU(s) & = \log{O_L} + \log{O_G}\\\nonumber
& = \log{\frac{P(S)}{P(X\cap Y \setminus S)}} + \log{\frac{P(S)}{1-P(S)}}\nonumber
\end{align}
As such, the decision on whether $a_x$ and $a_y$ should be merged to form $s$ can be made based solely on the \emph{Odds of Unithood (OU)} defined in Equation \ref{ou_3}. We will merge $a_x$ and $a_y$ if their odds of unithood exceeds a certain threshold, $OU_T$. 

\section{Evaluations and Discussions}\label{evaluation}

For this evaluation, we employed $500$ news articles from Reuters in the health domain gathered between December 2006 to May 2007. These $500$ articles are fed into the Stanford Parser whose output is then used by our head-driven left-right filter \cite{wong_2005,wong_et_al_2007b} to extract word sequences in the form of nouns and noun phrases. Pairs of word sequences (i.e. $a_x$ and $a_y$) located immediately next to each other, or separated by a preposition or the conjunction \emph{``and''} in the same sentence are measured for their unithood. Using the $500$ news articles, we managed to obtain $1,825$ pairs of words to be tested for unithood. 

We performed a comparative study of our new probabilistic approach against the empirically-derived unithood function described in Equation \ref{to_merge_or_not}. Two experiments were conducted. In the first one, we assessed our probabilistically-derived measure $OU(s)$ as described in Equation \ref{ou_3} where the decisions on whether or not to merge the $1,825$ pairs are done automatically. These decisions are known as the \emph{actual results}. At the same time, we inspected the same list manually to decide on the merging of all the pairs. These decisions are known as the \emph{ideal results}. The threshold $OU_T$ employed for our evaluation is determined empirically through experiments and is set to $-8.39$. However, since only one threshold is involved in deciding mergeability, training algorithms and data sets may be employed to automatically decide on an optimal number. This option is beyond the scope of this paper. The actual and ideal results for this first experiment are organised into a contingency table (not shown here) for identifying the true and the false positives, and the true and the false negatives. In the second experiment, we conducted the same assessment as carried out in the first one but the decisions to merge the $1,825$ pairs are based on the $UH(a_x,a_y)$ function described in Equation \ref{to_merge_or_not}. The thresholds required for this function are based on the values suggested by \cite{wong_et_al_2007b}, namely, $MI^+ = 0.9$, $MI^- = 0.02$, $ID_T = 6$, $IDR^+ = 1.35$, and $IDR^- = 0.93$.

\begin{figure}[htp]
    Table 1: The performance of $OU(s)$ (from Experiment 1) and $UH(a_x,a_y)$ (from Experiment 2) in terms of precision, recall and accuracy. The last column shows the difference in the performance of Experiment 1 and 2.\\
    \begin{center}
\includegraphics[width=2.8in]{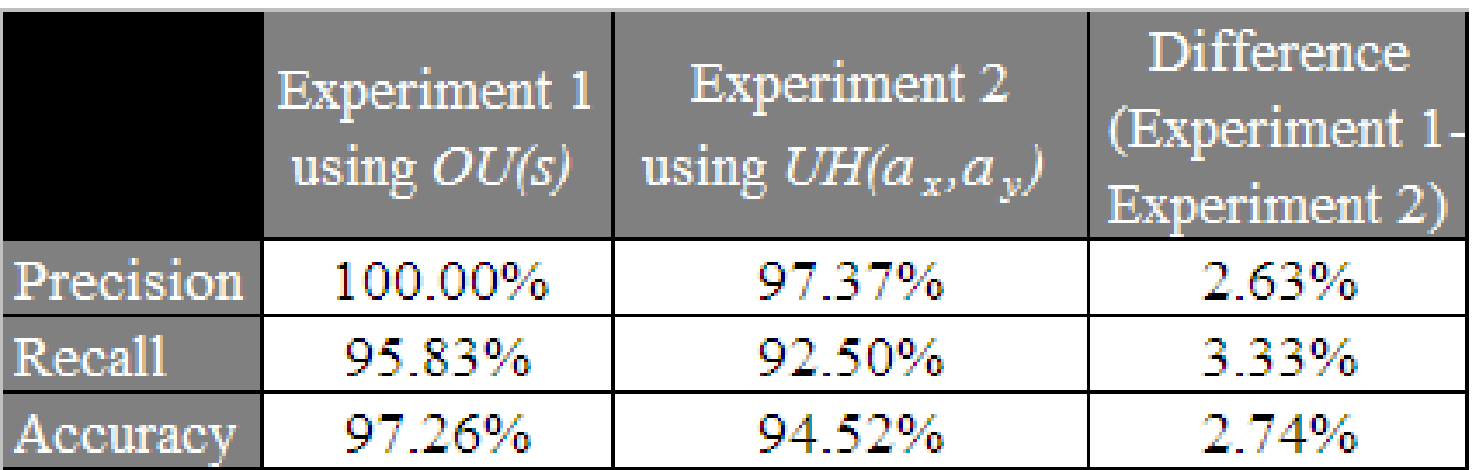}
    \end{center}
\end{figure}

Using the results from the contingency tables, we computed the precision, recall and accuracy for the two measures under evaluation. Table 1 summarises the performance of $OU(s)$ and $UH(a_x,a_y)$ in determining the unithood of $1,825$ pairs of lexical units. One will notice that our new measure $OU(s)$ outperformed the empirically-derived function $UH(a_x,a_y)$ in all aspects, with an improvement of $2.63\%$, $3.33\%$ and $2.74\%$ for precision, recall and accuracy, respectively. Our new measure achieved a $100\%$ precision with a lower recall at $95.83\%$. As with any measures that employ thresholds as a cut-off point in accepting or rejecting certain decisions, we can improve the recall of $OU(s)$ by decreasing the threshold $OU_T$. In this way, there will be less false negatives (i.e. pairs which are supposed to be merged but are not) and hence, increases the recall rate. Unfortunately, recall will improve at the expense of precision since the number of false positives will definitely increase from the existing $0$. Since our application (i.e. ontology learning) requires perfect precision in determining the unithood of word sequences, $OU(s)$ is the ideal candidate. Moreover, with only one threshold (i.e. $OU_T$) required in controlling the function of $OU(s)$, we are able to reduce the amount of time and effort spent on optimising our results. 

\section{Conclusion and Future Work}\label{conclusion}

In this paper, we highlighted the significance of unithood and that its measurement should be given equal attention by researchers in term extraction. We focused on the development of a new approach that is independent of influences of termhood measurement. We proposed a new probabilistically-derived measure which provide a dedicated way to determine the unithood of word sequences. We refer to this measure as the \emph{Odds of Unithood (OU)}. $OU$ is derived using Bayes Theorem and is founded upon two evidences, namely, \emph{local occurrence} and \emph{global occurrence}. Elementary probabilities estimated using page counts from the Google search engine are utilised to quantify the two evidences. The new probabilistically-derived measure $OU$ is then evaluated against an existing empirical function known as \emph{Unithood (UH)}. Our new measure $OU$ achieved a precision and a recall of $100\%$ and $95.83\%$ respectively, with an accuracy at $97.26\%$ in measuring the unithood of $1,825$ test cases. $OU$ outperformed $UH$ by $2.63\%$, $3.33\%$ and $2.74\%$ in terms of precision, recall and accuracy, respectively. Moreover, our new measure requires only one threshold, as compared to five in $UH$ to control the mergeability decision. 

More work is required to establish the \emph{coverage} and the \emph{depth} of the World Wide Web with regards to the determination of unithood. While the Web has demonstrated reasonable strength in handling general news articles, we have yet to study its appropriateness in dealing with unithood determination for technical text (i.e. the depth of the Web). Similarly, it remains a question the extent to which the Web is able to satisfy the requirement of unithood determination for a wider range of genres (i.e. the coverage of the Web). Studies on the effect of noises (e.g. keyword spamming) and multiple word senses on unithood determination using the Web is another future research direction.

\section*{Acknowledgement}
This research was supported by the Australian Endeavour International Postgraduate Research Scholarship, and the Research Grant 2006 by the University of Western Australia.

\bibliographystyle{acl}
\bibliography{ou_ijcnlp}

\end{document}